\documentclass[runningheads]{llncs}
\usepackage[T1]{fontenc}
\usepackage{graphicx}
\usepackage{comment}
\usepackage[ruled]{algorithm2e}
\usepackage{adjustbox}
\usepackage{tabularx}
\usepackage{booktabs}
\usepackage{multirow}
\usepackage{subcaption}
\usepackage{amsmath,amssymb}
\usepackage{hyperref}
\usepackage{setspace}
\usepackage{xcolor}
\begin{document}
\title{CARTGen-IR: Synthetic Tabular Data Generation for Imbalanced Regression
}
\titlerunning{Synthetic Tabular Data Generation for Imbalanced Regression}
%
\author{António Pedro Pinheiro\inst{1,2}\orcidID{0009-0000-2205-6265} \and
\\
Rita P. Ribeiro\inst{1,2}\orcidID{0000-0002-6852-8077}}
\authorrunning{A. P. Pinheiro, R. P. Ribeiro}
%
\institute{University of Porto, 4169-007 Porto, Portugal \and
INESC TEC, 4200-465 Porto, Portugal}
\maketitle              
\begin{abstract}

Handling imbalanced target distributions in regression poses a persistent challenge, as the underrepresentation of relevant target values can significantly hinder model performance. Existing data-level solutions often adapt classification-oriented techniques, introducing arbitrary thresholds over the continuous target and leading to artificial and potentially misleading problem formulations. Deep generative models offer flexible sample synthesis but are computationally intensive and difficult to interpret. We propose a CART-based synthetic sampling method specifically designed for imbalanced regression on tabular data. The method integrates relevance- and density-guided sampling to address sparse target regions without thresholding, and employs a feature-driven tree structure to generate realistic tabular samples across heterogeneous features and non-linear interactions. Experiments on benchmark datasets for extreme-value prediction show that the proposed approach is competitive with state-of-the-art resampling and generative methods while offering faster execution and greater transparency. These results highlight its potential as a scalable and interpretable data-level strategy for improving regression models in imbalanced domains.

\end{abstract}

\section{Introduction}
\label{sec:Intro}

Research on imbalanced domain learning for tabular data has predominantly centered on classification tasks~\cite{Ribeiro2020,10.1145/2907070}, with a strong emphasis on improving predictions for minority classes. Yet, many real-world problems involve continuous targets, where similar challenges arise in regression settings. These include forecasting extreme weather events~\cite{schultz2021}, predicting exceptionally high sea surface temperatures~\cite{ALERSKANS2022113220}, identifying unusually low drug responses in cancer cells that signal increased sensitivity~\cite{lenhof2022}, and detecting major financial frauds. Unlike classification, regression tasks introduce an additional difficulty: determining which regions of the continuous target space should be considered rare, relevant, and worthy of focused modeling.


Data-level approaches remain among the most widely used strategies for addressing imbalance: they modify the training distribution to emphasize the most relevant cases, offer considerable flexibility, and can be applied during preprocessing, thereby extending the applicability of standard machine learning algorithms~\cite{10.1145/2907070}.

In imbalanced regression, several data-level techniques have been proposed \cite{ribeiro2019,ribeiro2013,Branco2017,camacho2022,Camacho2024,islam2024,TIAN2023119157,pr12020375,stocksieker2024vae}. However, many of these proposals exhibit intrinsic limitations. Methods that discretize the continuous target variable through artificial thresholds often compromise interpretability and transparency, effectively transforming the approach into a black-box. Moreover, overly simplistic data-generation mechanisms can increase the risk of overfitting, and numerous strategies have limited ability to handle categorical features or missing values.

To address these issues, this paper adapts a CART-based data augmentation method~\cite{panagiotou2024synthetictabulardatageneration}, originally designed for classification, to the context of imbalanced regression. Our approach avoids user-defined thresholds on the target variable, thereby eliminating the arbitrariness associated with domain discretization in existing methods~\cite{Branco2017,ribeiro2015,ribeiro2019}. Moreover, because the technique leverages decision trees for synthetic data generation, it inherits the transparency and interpretability characteristic of CART models, while naturally supporting numeric and categorical features as well as missing values.

The remainder of this paper is structured as follows: Section~\ref{sec:RelatedWork} delineates the problem definition and provides a concise overview of the pertinent literature, including existing oversampling and data augmentation techniques. Section~\ref{sec:Proposal} presents our proposed solutions, while Section~\ref{sec:Experiments} discusses the outcomes of an extensive experimental evaluation. Concluding remarks are provided in Section~\ref{sec:Concl}.

\section{Related Work}
\label{sec:RelatedWork}


Supervised learning aims to approximate an unknown function \(Y = f(X_1, \dots, X_p)\) using a model \(h\) trained on data \(D = \{\langle \mathrm{x}_i, y_i\rangle\}_{i=1}^n\), with tasks framed as classification or regression depending on 
the nature of target variable 
\(Y\), 
and optimized through criteria such as error rate or squared error~\cite{10.1145/2907070,Ribeiro2020}. 
In many regression tasks, the most relevant cases correspond to extreme target values. Their underrepresentation biases models toward predicting average outcomes, causing them to overlook rare yet critical extremes. 
Addressing such imbalanced regression problems~\cite{10.1145/2907070,Ribeiro2011} involves (i) identifying key target ranges, (ii) assessing performance in these areas, and (iii) biasing learning toward rare instances~\cite{Ribeiro2020}. Existing research tackles these issues from different perspectives~\cite{Ribeiro2020,Branco2017,ribeiro2015,ribeiro2019,camacho2022,ribeiro2013,Camacho2024,islam2024,TIAN2023119157,pr12020375}. 
This paper focuses on (i), adopts suitable metrics for (ii), by selecting specially suited metrics for imbalanced regression tasks, instead of commonly used standard error metrics, and proposes a new data-level method for (iii).


A central challenge in imbalanced regression is defining non-uniform preferences over continuous domains. While full domain knowledge would ideally guide this process, according to the user needs and preferences, it is rarely available, particularly for infinite target domains. Two main approaches address this issue. The first, proposed by~\cite{Ribeiro2011,Ribeiro2020}, approximates a relevance function \(\phi()\in[0,1]\) via interpolation of domain-based control points; when knowledge is lacking, it employs a non-parametric, data-driven procedure that assumes extreme values are most important and derives control points from adjusted boxplot statistics~\cite{HUBERT20085186}. The second, DenseWeight~\cite{Steininger2021}, assigns weights inversely proportional to the estimated density of target values using Kernel Density Estimation (KDE), emphasizing underrepresented outcomes by prioritizing low-density regions; all weights are positive and normalized to ensure stable gradient descent. 


Data-level strategies address imbalanced regression by modifying data distributions to better represent relevant instances, using sampling and data augmentation techniques~\cite{10.1145/2907070}. Examples include Random Undersampling (RU), Random Oversampling (RO), and WEighted Relevance Combination Strategy (WERCS) \cite{ribeiro2019}, which employ relevance functions for guided sampling. Gaussian Noise (GN) \cite{ribeiro2019} augments data by perturbing relevant cases while undersampling common ones. Additionally, SMOTER~\cite{ribeiro2013} adapts SMOTE for regression by interpolating synthetic samples based on relevance, while SMOGN~\cite{Branco2017} combines undersampling with SMOTER and GN to balance synthetic data fidelity and variability. Furthermore, G-SMOTER~\cite{camacho2022} extends G-SMOTE using geometric transformations to diversify generated samples. 

However, many of these methods stem from classification settings, often relying on arbitrary thresholds to partition continuous target spaces, which is ill-suited for regression tasks due to the introduction of artificial discretization in the target domain. This is particularly detrimental to the nature of a continuous target variable. By introducing a threshold, the user, for instance, stipulates that any value exceeding the threshold is significant and valuable, while any sample below the threshold is deemed common and insignificant. 
Consequently, if the threshold is, for example, $x = 5$, then $5.1$ is considered important, whereas $4.9$ is not. Intuitively, this does not make much sense and creates an abrupt division within the domain. Furthermore, it implies that, for instance, a sample with a target value of $1.2$ and another sample with $4.9$ are both categorized as normal, common cases, but the magnitude of the target variable value is disregarded. Consequently, artificial thresholds end up categorizing an infinite domain and eliminating the inherent value of a continuous target variable.

WSMOTER~\cite{Camacho2024} mitigates this by integrating DenseWeight with SMOTE, using probabilistic weighting to focus on sparse regions. KNNOR-REG~\cite{islam2024} enhances SMOTE with k-NN filtering to identify representative minority points, addressing intra-domain imbalance and noise. Deep learning models like GANs and VAEs have also been explored for synthetic data generation,  with some especially developed in the context of imbalanced regression tasks. 
Models like TVAE~\cite{xu2019modelingtabulardatausing}, CTGAN~\cite{xu2019modelingtabulardatausing},CopulaGAN~\cite{patki2016synthetic} and diffusion-based TabDDPM~\cite{kotelnikov2023tabddpm} have been widely studied for synthetic case generation but lack specific focus on imbalanced regression tasks.
DIRVAE~\cite{TIAN2023119157} uses a dual-model GAN framework to improve generative performance on sparse regression data. IRGAN~\cite{pr12020375} integrates generation, correction, discrimination, and regression modules for synthetic sample creation. DAVID~\cite{stocksieker2024vae} combines regression training with a \(\beta\)-VAE architecture. Despite their flexibility, these generative models are computationally intensive and offer limited interpretability due to their black-box nature.

In \cite{reiter2005}, the author introduced the use of the CART algorithm~\cite{breiman1984} for synthetic data generation to protect sensitive microdata, emphasizing its advantages over parametric models in handling unknown distributions, complex interactions, and missing values without explicit imputation. Following the same idea, \cite{akiya2024} applied CART to generate synthetic patient data for survival analysis in oncology trials, where it outperformed methods such as Random Forests, Bayesian Networks, and CTGAN—particularly in low-data settings.
Building further on the ability of CART to model complex relationships, \cite{panagiotou2024synthetictabulardatageneration} evaluated a CART-based approach for synthetic tabular data generation in the context of class imbalance and fairness, showing strong performance compared with traditional techniques such as SMOTE and GANs. Overall, prior research demonstrates that CART-based data generation effectively captures complex dependencies in the original data while avoiding the distortions often introduced by SMOTE and GANs. Further expanding on these tree-based mechanisms,\cite{caiola2010randomforestsforgenerating} proposes a random forest approach for regression augmentation. By sampling from the terminal nodes of the ensemble, they approximate the underlying conditional distributions and effectively model non-linear dependencies.

\section{
Our Proposal
}
\label{sec:Proposal}




Building on the good results of CART-based synthetic data generation reported in the literature~\cite{reiter2005,akiya2024,panagiotou2024synthetictabulardatageneration}, we propose CARTGen-IR, a method that uses CART to generate synthetic tabular data for imbalanced regression. Our approach removes the need for arbitrary user-defined thresholds when identifying relevant or rare cases, an issue that conflicts with the continuous nature of regression targets. By avoiding crisp partitions of the target domain, CARTGen-IR prevents the domain discretization required by methods such as SMOTER~\cite{ribeiro2013} and SMOGN~\cite{Branco2017}.
Since the method builds on decision trees, it samples according to the conditional distributions estimated by CART while also retaining the algorithm’s inherent transparency and interpretability.
The recursive partitioning process provides an auditable view of how synthetic data is generated. In addition, CARTGen-IR naturally handles numeric and categorical variables, as well as missing values, making it broadly applicable to tabular data.

The overall procedure of CARTGen-IR is summarized in Algorithm \ref{alg:cart-ir}. The method begins by weighting target values according to their rarity or relevance, assigning higher weights to rare cases. Rarity is estimated using either the DenseWeight method~\cite{Steininger2021} or the relevance function~\cite{Ribeiro2020}, selected through the $\rho$ weight scheme hyperparameter. The former uses probabilistic estimation, which may not match user preferences when density imbalance doesn’t reflect their priorities, unlike the relevance-based approach. These scores are adjusted with a rarity exponent~$\alpha$, normalized, and used to resample the dataset with replacement so that rare instances are more likely to be selected.

Rarity-based weighting is also employed in methods such as WERCS~\cite{ribeiro2019} and G-SMOTER~\cite{camacho2022}, but CARTGen-IR distinguishes itself by resampling the original dataset in a way that largely excludes frequent target values from augmentation. This produces synthetic data that concentrates more effectively on the rare and relevant regions of the target space. Although we implement DenseWeight and the relevance function, the method supports any rarity estimation mechanism without requiring changes to the overall framework, ensuring adaptability to future developments.

The extent of resampling is controlled by the hyperparameter~$\eta \in [0,1]$, which determines the proportion of synthetic samples to be generated. 
Following established practice in synthetic data generation~\cite{Chawla_2002,he2008adasyn}, each selected instance serves as the basis for generating multiple synthetic samples.

\begin{figure}[hbt]
    \centering
    \includegraphics[width=1\linewidth]{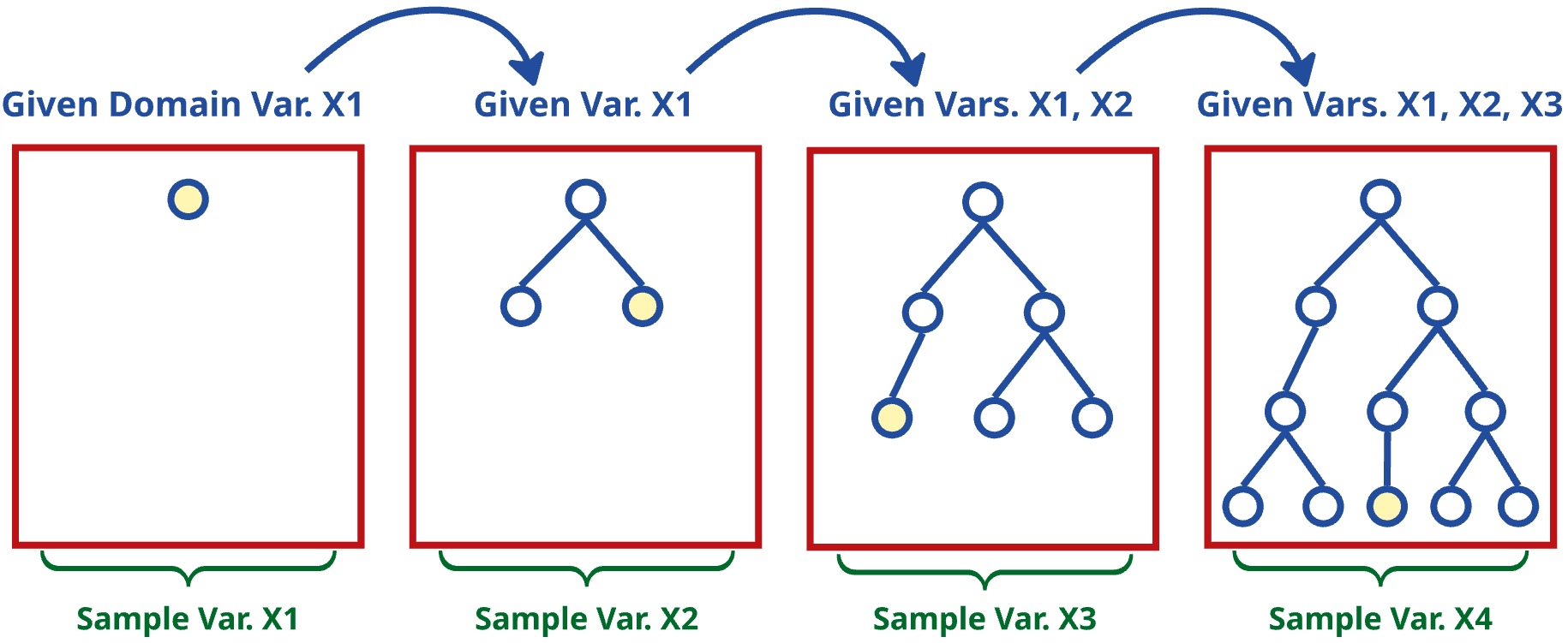}
    \caption{CART consecutive column-wise data generation.}
    \label{fig:cart_data_generation}
\end{figure}

Because resampling can create duplicate instances, an optional noise mechanism controlled by~$\delta$ can be applied to numeric features to reduce overfitting. The resampled dataset, enriched with rare cases, is then used to generate synthetic data through a CART-based, sequential, attribute-wise data generation procedure via sampling in the leaves, described in~\cite{panagiotou2024synthetictabulardatageneration,reiter2005}, where a CART model is iteratively fitted for each attribute using the remaining, previous attributes as predictors. The synthesis follows a three-step protocol. First, 
several CART trees are trained by considering each variable as the target and using only the preceding variables as predictors (e.g., the model for $X_3$ is trained using $X_1$ and $X_2$), through the \texttt{FitCARTModels} function. Second, the \texttt{GenSynthetic} function generates data for a new synthetic case by iteratively selecting the appropriate terminal/leaf node for each variable based on the values of the preceding variables already generated for that case. For the first variable in the sequence, where no prior values exist for the synthetic sample,  the process defaults to a root node, and the entire pool of values for that variable is used to select a random value. Third, a value is then randomly drawn from the members of that node and the observed value is taken as the synthetic output for that variable. Furthermore, for continuous variables, instead of sampling directly from the discrete values in a leaf node, the method fits a Gaussian kernel density estimator to the values in that leaf and samples from that smooth distribution. This improves generalization by generating plausible values that fit the statistical profile, including sparse tails, without being limited to the exact points found in the training set. This process is repeated for each synthetic case, always using the same trees fitted in the \texttt{FitCARTModels} stage, derived solely from the original data, and is not affected by the new synthetic data generated. This process is illustrated in Figure~\ref{fig:cart_data_generation}.

\begin{algorithm}[t]
\caption{CARTGen-IR Algorithm}
\label{alg:cart-ir}
\scriptsize
\begin{spacing}{1.1}
\DontPrintSemicolon
\SetKwFunction{FMain}{CARTGen-IR}
\SetKwProg{Fn}{function}{:}{}
\Fn{\FMain{$\langle\mathbf{X},Y\rangle$,
$\rho$: \textnormal{weight scheme}, 
$\alpha$: \textnormal{exponent}, 
$\eta$: 
\textnormal{sampling prop.}, $\delta$: \textnormal{noise}}}{

    $\mathbf{w} \gets$ \texttt{RarityWeights}($Y$, $\rho$, $\alpha$) \\

    $N \gets \lfloor \eta \cdot |Y| \rfloor$ \tcp*{\tiny \textnormal{nr. cases to generate}} 
    $N_{re} \gets \lfloor N / 5 \rfloor$ \tcp*{\tiny \textnormal{nr. cases to resample~\cite{Chawla_2002,he2008adasyn}}} 
$\langle\mathbf{X}_{\text{new}},Y_{\text{new}}\rangle \gets$ 
    \texttt{sample}($\langle\mathbf{X},Y\rangle,N_{re},\mathbf{w},repl= \texttt{T}$) \\
    \uIf{$\delta > 0$}{$\mathbf{X}_{\text{new}} \gets \texttt{JitterDuplicates}(\mathbf{X}_{\text{new}}, \delta)$%
    \tcp*{\tiny \textnormal{add Gaussian noise to duplicates}}}    
    Trees $\gets$ \texttt{FitCARTModels}($\langle\mathbf{X}_{\text{new}},Y_{\text{new}}\rangle$)
\tcp*{\tiny \textnormal{one CART tree per attribute as target}}
    $\langle X_{\text{synth}},Y_{\text{synth}}\rangle \gets$ \texttt{GenSynthetic}(Trees, $N$) \tcp*{\tiny \textnormal{attribute-wise generation of synthetic data}}

    \KwRet{$\langle \mathbf{X},Y \rangle \cup \langle\mathbf{X}_{\text{synth}},Y_{\text{synth}}\rangle$}
}

\SetKwFunction{FRarityWeights}{RarityWeights}
\SetKwProg{FnRarity}{function}{:}{}
\FnRarity{\FRarityWeights{$Y$, $\rho$, $\alpha$}}{
    \uIf{$\rho = $ \emph{denseweight}}{
        $\mathbf{w} \gets $\texttt{DenseWeight}$(Y)$ \tcp*{\tiny \textnormal{ algorithm presented in \protect\cite{Steininger2021}}}
    }
    \uElseIf{$\rho = $ \emph{relevance}}{
        $\mathbf{w} \gets \phi(Y)$ \tcp*{\tiny \textnormal{algorithm presented in \protect\cite{Ribeiro2020}}}
    }
    $\mathbf{w} \gets \mathbf{w}^\alpha$ \\
    \KwRet{$\mathbf{w}/\sum \mathbf{w}$}
}
\end{spacing}
\end{algorithm}

\begin{figure}[htbp]
    \centering
    \begin{subfigure}[b]{0.47\textwidth}
        \centering
        \includegraphics[width=\linewidth]{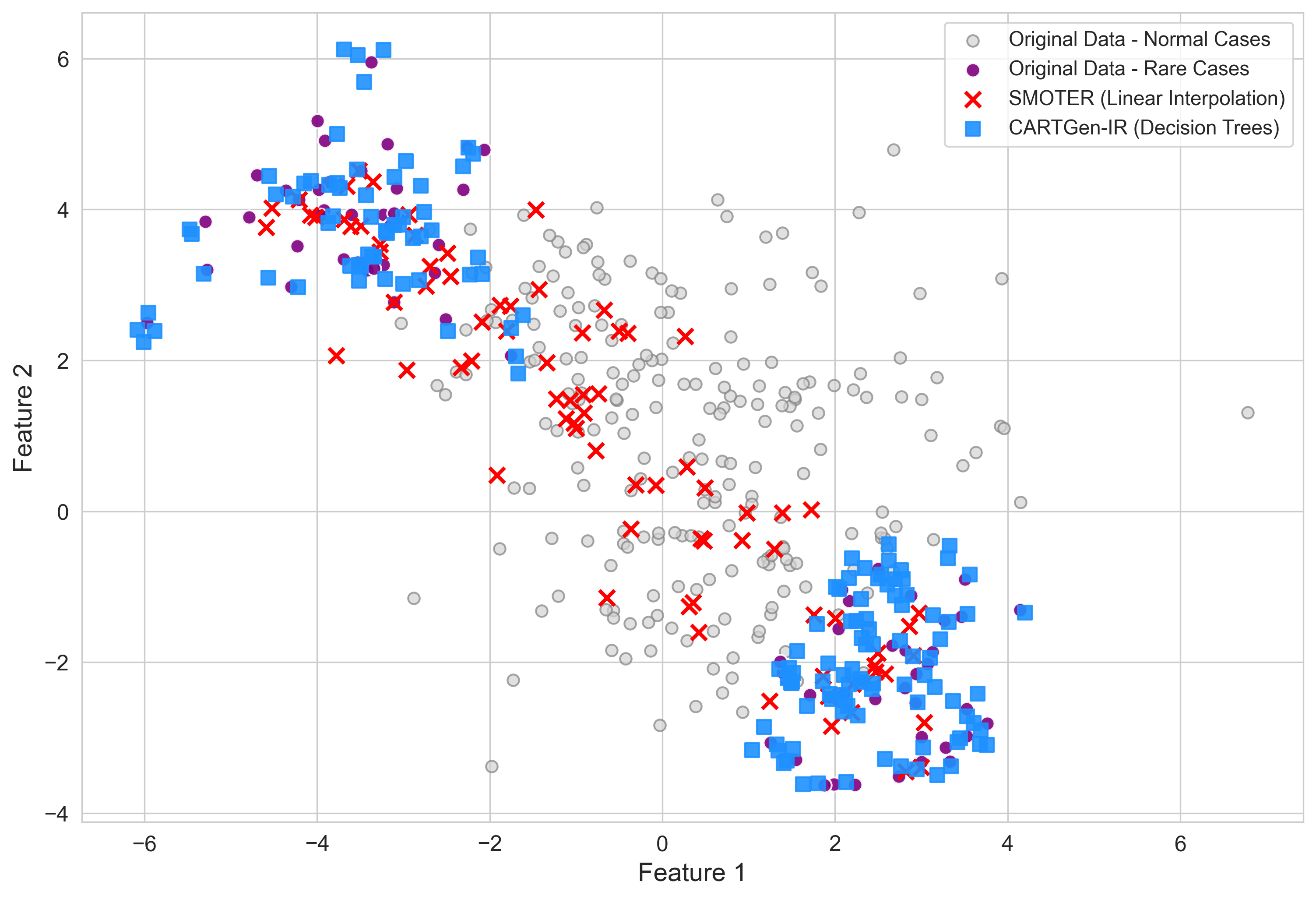}
        \caption{\scriptsize CARTGen-IR vs. SMOTER}
        \label{fig:cartgen_vs_smoter}
    \end{subfigure}
    \hfill 
    \begin{subfigure}[b]{0.47\textwidth}
        \centering
        \includegraphics[width=\linewidth]{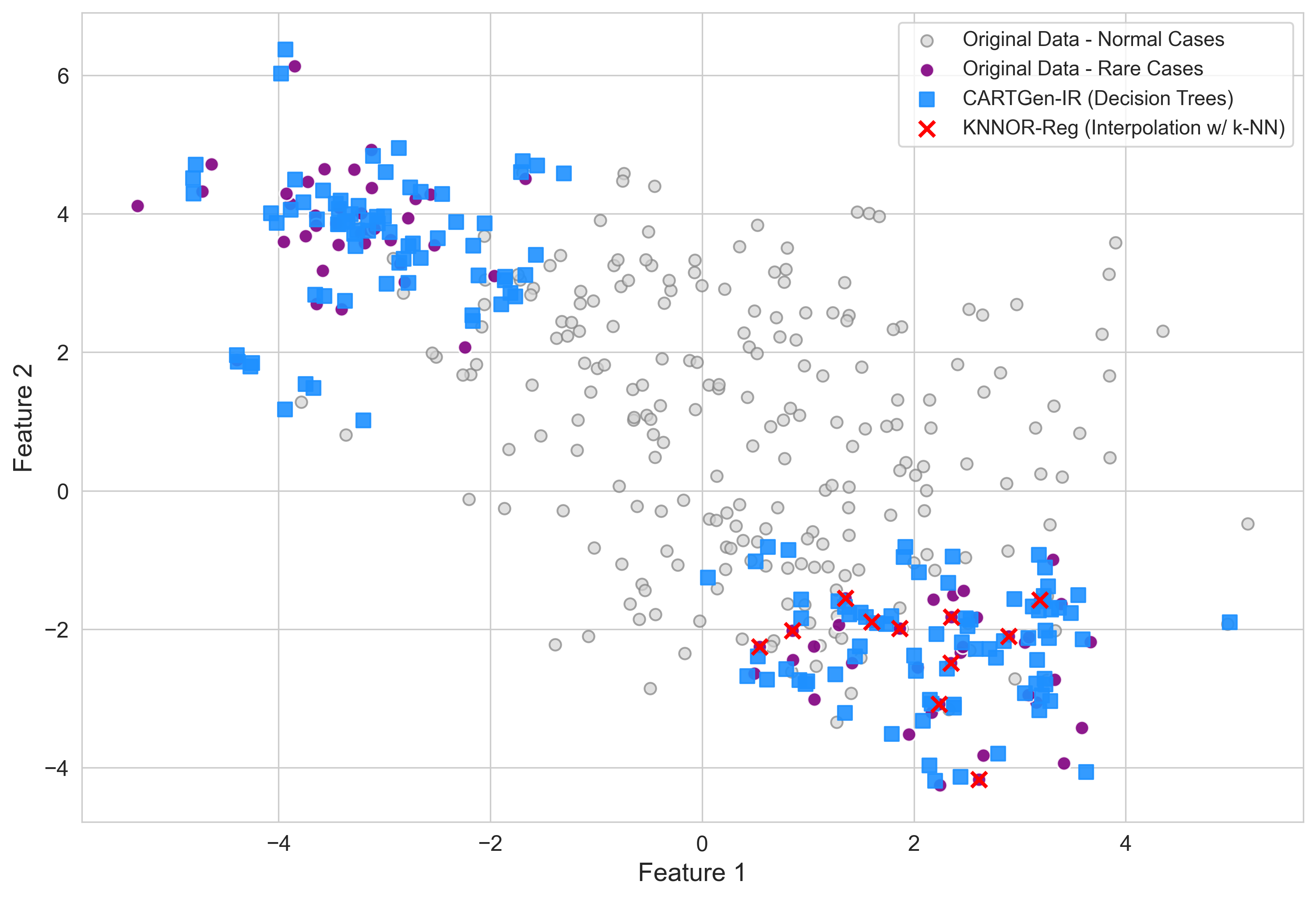} 
        \caption{\scriptsize CARTGen-IR vs. KNNOR-REG}
        \label{fig:cartge_vs_knnorreg}
    \end{subfigure}
    
    \caption{Synthetic sample generation comparison.}
    \label{fig:sample_generation_cartgen_ir_smoter}
\end{figure}

Figure~\ref{fig:sample_generation_cartgen_ir_smoter} shows a simple example with two features and target values labeled as common or rare. This setup shows how CARTGen-IR handles imbalanced regression differently from interpolation-based methods. While SMOTER creates unrealistic synthetic points and KNNOR-REG fails to generate diverse feature values and only performs augmentation in one rare cluster, CARTGen-IR uses tree-based partitioning to generate samples locally within rare regions, preserving the original data structure.

\section{Experimental Study}
\label{sec:Experiments}

Our primary objective is to evaluate CARTGen-IR performance and capabilities within the broader context of a comparative study against other state-of-the-art data-level strategies for imbalanced regression tasks.
With the experimental study, we aim to answer the following research questions: (\textbf{RQ1}) Is CARTGen-IR effective for imbalanced regression scenarios, both in terms of standard error metrics and specially-suited error metrics? 
(\textbf{RQ2}) How does it compare to the state-of-the-art data-level methods proposed for imbalanced regression tasks for tabular data, in both rare and common target subdomains?; and, finally (\textbf{RQ3}) What tradeoff do these methods offer regarding predictive performance and execution time?


\subsection{Experimental Setup}
\label{sec:ExpSetup}

In this study, we used 15 widely used regression datasets. The key attributes of these datasets are summarized in Table \ref{tab:datasets}, which also reports the absolute and relative frequencies of rare instances, as well as the types of extremes, defined according to a relevance threshold of 0.8. To accomplish this, we derived a relevance function for each dataset using the automated approach outlined in~\cite{Ribeiro2020}. The chosen datasets present a broad range of characteristics such as numeric and nominal features, instances, types of extremes, and rare occurrences.

\begin{table}[t]
\centering
\caption{
Datasets used in the experiments, with nr. of cases, nominal and numerical features, extreme-value type, and rarity statistics for a threshold of 0.8.
}
\renewcommand{\arraystretch}{0.9}
\label{tab:datasets}
\small
\begin{tabular}{
lrccccc
}
\toprule
\textbf{Dataset} & \textbf{\#cases} &  \textbf{\#NomFeat} & \textbf{\#NumFeat} & \textbf{ExtrType} & \textbf{\# Rare} & \textbf{\% Rare} \\
\midrule
strikes         & 625  & 0  & 6  & High & 15   & 2.40 \\
forestFires     & 517  & 0  & 12 & High & 15   & 2.90 \\
ele-1           & 495  & 0  & 2  & High & 21   & 4.24 \\
cpuSm           & 8,192 & 0  & 12 & Low  & 371  & 4.53 \\
airfoil         & 1,503 & 0  & 5  & High & 80   & 5.32 \\
fuelConsumption & 1,764 & 12 & 25 & Both & 167  & 9.47 \\
heat            & 7,400 & 3  & 8  & Both & 833  & 11.26 \\
sensory         & 576  & 0  & 11 & Both & 69   & 11.98 \\
mortgage        & 1,049 & 0  & 15 & Low  & 133  & 12.68 \\
maxTorque       & 1,802 & 13 & 19 & Both & 235  & 13.04 \\
treasury        & 1,049 & 0  & 15 & Low  & 137  & 13.06 \\
availablePower  & 1,802 & 7  & 8  & Both & 305  & 16.93 \\
housingBoston   & 506  & 0  & 13 & Both & 105  & 20.75 \\
abalone         & 4,177 & 1  & 7  & Both & 1033 & 24.73 \\
servo           & 167  & 4  & 0  & Both & 59   & 35.33 \\
\bottomrule
\end{tabular}
\end{table}

We evaluated a comprehensive set of preprocessing strategies to address data imbalance in regression tasks. The approaches considered include
RU, RO, WERCS and GN from \cite{ribeiro2019}, SMOTER~\cite{ribeiro2013}, SMOGN~\cite{Branco2017}, WSMOTER~\cite{Camacho2024}, G-SMOTER~\cite{camacho2022}, DAVID~\cite{stocksieker2024vae}, KNNOR-REG~\cite{islam2024} and CARTGen-IR, which were 
specifically developed for imbalanced regression tasks. Furthermore, we included other state-of-the-art deep learning techniques which, although not developed purposefully to tackle imbalanced regression problems, are still suitable for these scenarios: (i) VAE-based method: TVAE~\cite{xu2019modelingtabulardatausing}, (ii) GAN-based methods: CTGAN~\cite{xu2019modelingtabulardatausing} and CopulaGAN~\cite{patki2016synthetic}; 3) diffusion-based method: TabDDPM~\cite{kotelnikov2023tabddpm}.
A complete overview of the 56 resampling configurations is presented in Table~\ref{tab:resampling_strategies}. The hyperparameter values used for each preprocessing strategy were selected based on those used in the literature where they were proposed or introduced. For CARTGen-IR, the hyperparameter search space was defined by synthesizing established values from related literature and refining them through preliminary empirical analysis. It is important to note that a direct comparison with the DIRVAE approach proposed by~\cite{TIAN2023119157} as well as the IRGAN approach proposed by~\cite{pr12020375} was not possible, as both source codes are not publicly available.

\begin{table}[t]
\caption{Data-level strategies and their hyperparameter settings}
\label{tab:resampling_strategies}
\small
\centering
\setlength{\tabcolsep}{1.2mm}
\resizebox{\textwidth}{!}{
\begin{tabular}{ll
}
\toprule
\textbf{Strategy} & \textbf{Hyperparameters} \\
\midrule
RU~\cite{ribeiro2019} & \%u = \{balance, extreme\} \\
RO~\cite{ribeiro2019} & \%o = \{balance, extreme\} \\
WERCS~\cite{ribeiro2019} & \%u/\%o = \{0.5, 0.75\} \\
GN~\cite{ribeiro2019} & \%u/\%o = \{balance, extreme\}, $\delta$ = \{0.05, 0.1, 0.5\} \\
SMOTER~\cite{ribeiro2013} & \%u/\%o = \{balance, extreme\} \\
SMOGN~\cite{Branco2017} & \%u/\%o = \{balance, extreme\}, $\delta$ = \{0.05, 0.1, 0.5\} \\
WSMOTER~\cite{Camacho2024} & ratio = \{1.5, 1.75\}, $\beta$ = \{1, 2\} \\
G-SMOTER~\cite{camacho2022} & strategy = \{minority, majority, combined\}, $\xi$ = 0.7, $\eta$ = 0.75, $\theta$ = \{-0.5, 0.5\}, $k$ = 5 \\
DAVID~\cite{stocksieker2024vae} & $\alpha$ = \{1, 2\} \\
TVAE~\cite{xu2019modelingtabulardatausing} & -- \\
CTGAN~\cite{xu2019modelingtabulardatausing} & -- \\
CopulaGAN~\cite{patki2016synthetic} & -- \\
TabDDPM~\cite{kotelnikov2023tabddpm} & -- \\
KNNOR-REG~\cite{islam2024} & -- \\
CARTGen-IR & $\rho$ = \{denseweight, relevance\}, $\alpha$ = \{1.0, 1.5, 2.0\}, $\eta$ = \{0.5, 0.75\},  $\delta$ = \{0.001, 0\} \\
\bottomrule
\end{tabular}
}
\end{table}

To assess the effectiveness of these data-level strategies, we paired them with three algorithms: Random Forest (RF), Support Vector Regressor (SVR), and XGBoost (XGB).
The experimental setup had 14 hyperparameter combinations for tuning the learning models: RF with ’n\_estimators’ (100, 200) and ‘max\_features’ (‘sqrt’, ‘log2’), SVR with ‘rbf’ kernel, ‘C’ (1, 10, 100), and ‘epsilon’ (0.1, 0.5), and XGB with ’n\_estimators’ (100, 200) and ‘max\_depth’ (3, 6).
Each model was evaluated across 15 regression datasets under all 56 preprocessing conditions, resulting in a total of 11.760 experiments ($14 \times 15 \times 56$). 

In this study, we used both the Squared Error–Relevance Area (SERA)~\cite{Ribeiro2020} and the relevance weighted RMSE (RW-RMSE), proposed by~\cite{imbalance-metrics} as evaluation metrics to better assess model performance under imbalanced conditions. We also included RMSE as a standard regression error metric. 

All evaluation metrics were computed using a stratified, repeated 2×5-fold cross-validation procedure to ensure robust and reliable performance estimates, coupled with a nested grid search to simultaneously tune the regressor hyperparameters and the sampling strategy configurations. Data augmentation was applied strictly within the cross-validation loop on the training folds only, ensuring no synthetic data leaked to the test sets. The stratification procedure was conducted considering the quantiles of the target variable.

\subsection{Obtained Results}
\label{sec:results}

\begin{figure}[ht]
    \centering
\resizebox{\textwidth}{!}{
\includegraphics{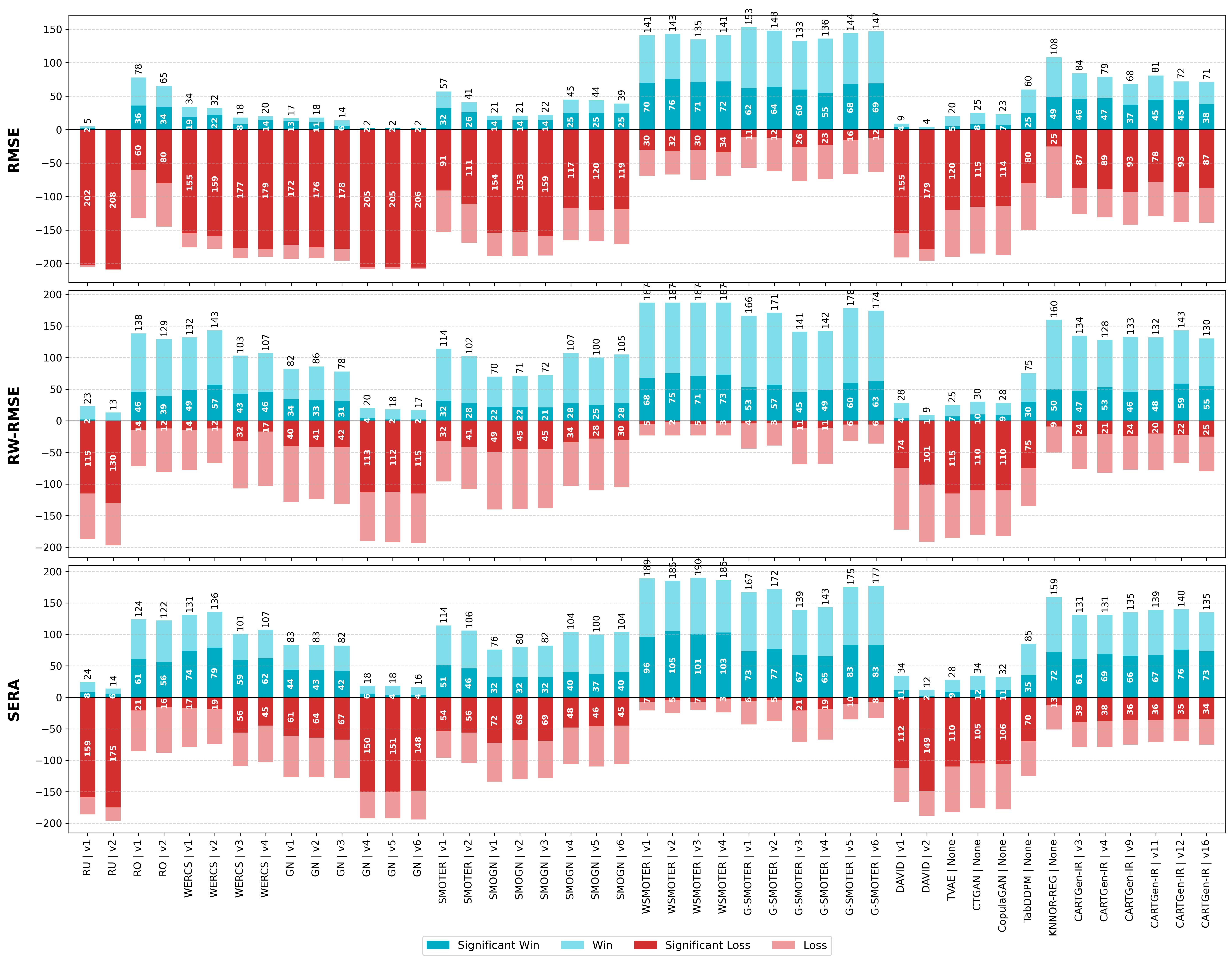}
}
    \caption{Comparison of wins and losses, including significant outcomes at the 95\% confidence level, for each data-level strategy against a no-preprocessing baseline, across different learners, datasets, and evaluation metrics.
    }
    \label{fig:wins_losses}
\end{figure}

Figure~\ref{fig:wins_losses} provides an overview of the wins and losses of each data augmentation method across all evaluation metrics, highlighting statistically significant differences determined using the Wilcoxon signed-rank test at the 95\% confidence level. For readability, the figure reports only the six best-performing configurations of CARTGen-IR (out of 16 tested), together with all competing approaches. The purpose of this test is to enable a comparative analysis among the methods without an absolute pre-determined threshold for distinguishing between good and poor performances.


From this analysis, WSMOTER emerges as the most consistent performer across metrics, followed by KNNOR-REG and G-SMOTER. CARTGen-IR ranks fourth in overall consistency, with a level of robustness not observed in other methods. For instance, WERCS performs well in its oversampling-dominant versions but declines when undersampling becomes more prominent. Notably, although CARTGen-IR is not the most frequent winner, it exhibits a superior significant win-to-loss ratio compared to similarly performing methods, indicating that its victories are generally more meaningful. 
Additionally, the strongest CARTGen-IR variants exhibit a consistent set of characteristics: they introduce Gaussian noise to numerical attributes and perform well under both density-weighting schemes and rarity exponents of 1.5 and 2.0. The proportion of synthetic samples has little influence, as both tested values produced similar results.

\begin{figure}[!ht]
    \centering
    
    \begin{subfigure}[t]{0.32\textwidth}
        \centering
        \includegraphics[width=1\linewidth]{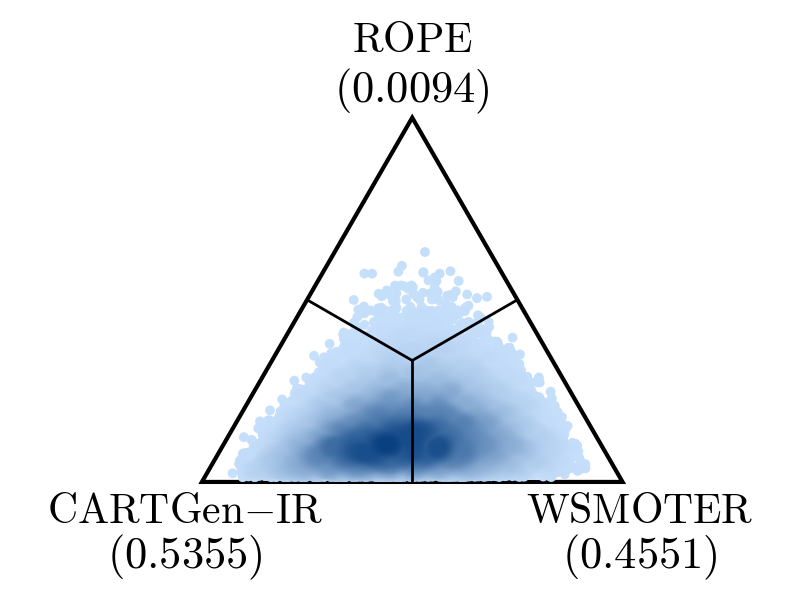}
        \caption{RMSE – SVR}
        \label{fig:main_rmse_svr_triangle_bayesian}
    \end{subfigure}
    \hspace{0.00\textwidth}
    \begin{subfigure}[t]{0.32\textwidth}
        \centering
        \includegraphics[width=1\linewidth]{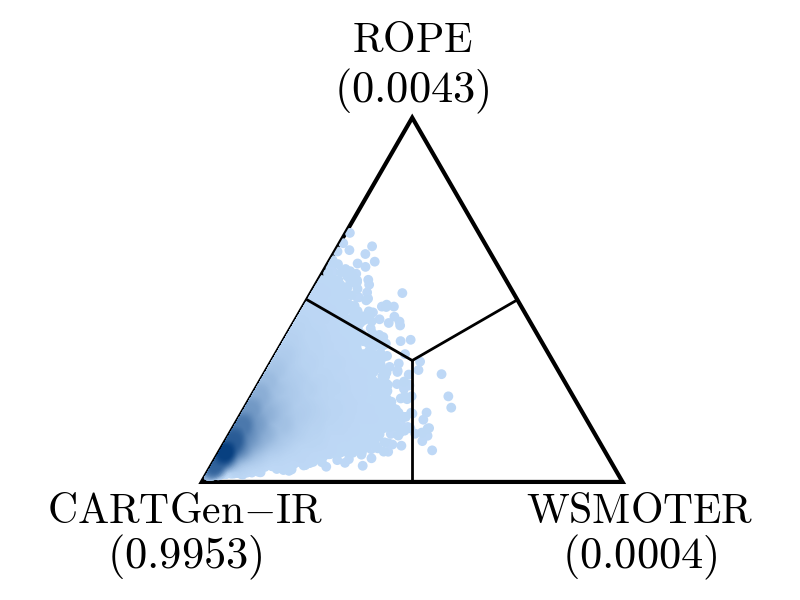}
        \caption{RW-RMSE – RF}
        \label{fig:main_wrmse_rf_triangle_bayesian}
    \end{subfigure}
    \hspace{0.00\textwidth}
    \begin{subfigure}[t]{0.32\textwidth}
        \centering
        \includegraphics[width=1\linewidth]{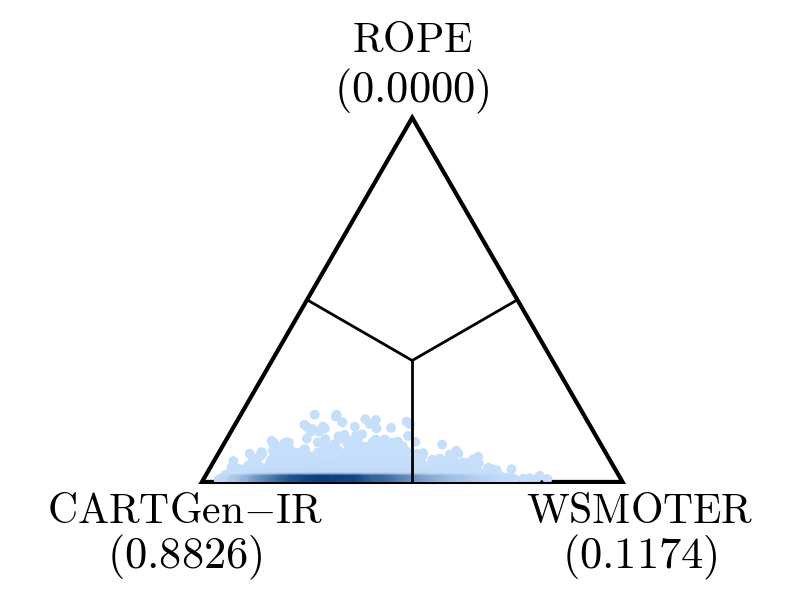}
        \caption{SERA – XGBoost}
        \label{fig:main_sera_xgboost_triangle_bayesian}
    \end{subfigure}

    \caption{Bayesian posterior ternary plots.}
    \label{fig:main_ternary_bayesian}
\end{figure}

Following the observation of strong performance from both CARTGen-IR and WSMOTER from previous tests, a Bayesian signed-rank test was conducted to compare them, utilizing a Region of Practical Equivalence (ROPE) of $[-1\%, 1\%]$. The results indicated that CARTGen-IR consistently outperformed WSMOTER across the evaluated models. Specifically, for RF, CARTGen-IR demonstrated a 99\% probability of superiority across all metrics - Figure~\ref{fig:main_wrmse_rf_triangle_bayesian}. While SVR performance was balanced regarding RMSE and RW-RMSE (Figure~\ref{fig:main_rmse_svr_triangle_bayesian}), CARTGen-IR dominated the SERA metric with probabilities exceeding 90\%. For XGBoost, CARTGen-IR showed a probability of superiority above 70\% for most metrics - Figure~\ref{fig:main_sera_xgboost_triangle_bayesian}. Overall, the analysis suggests that CARTGen-IR is generally the superior method.

\begin{figure}[ht]
    \centering
    \resizebox{\textwidth}{!}{
    \includegraphics{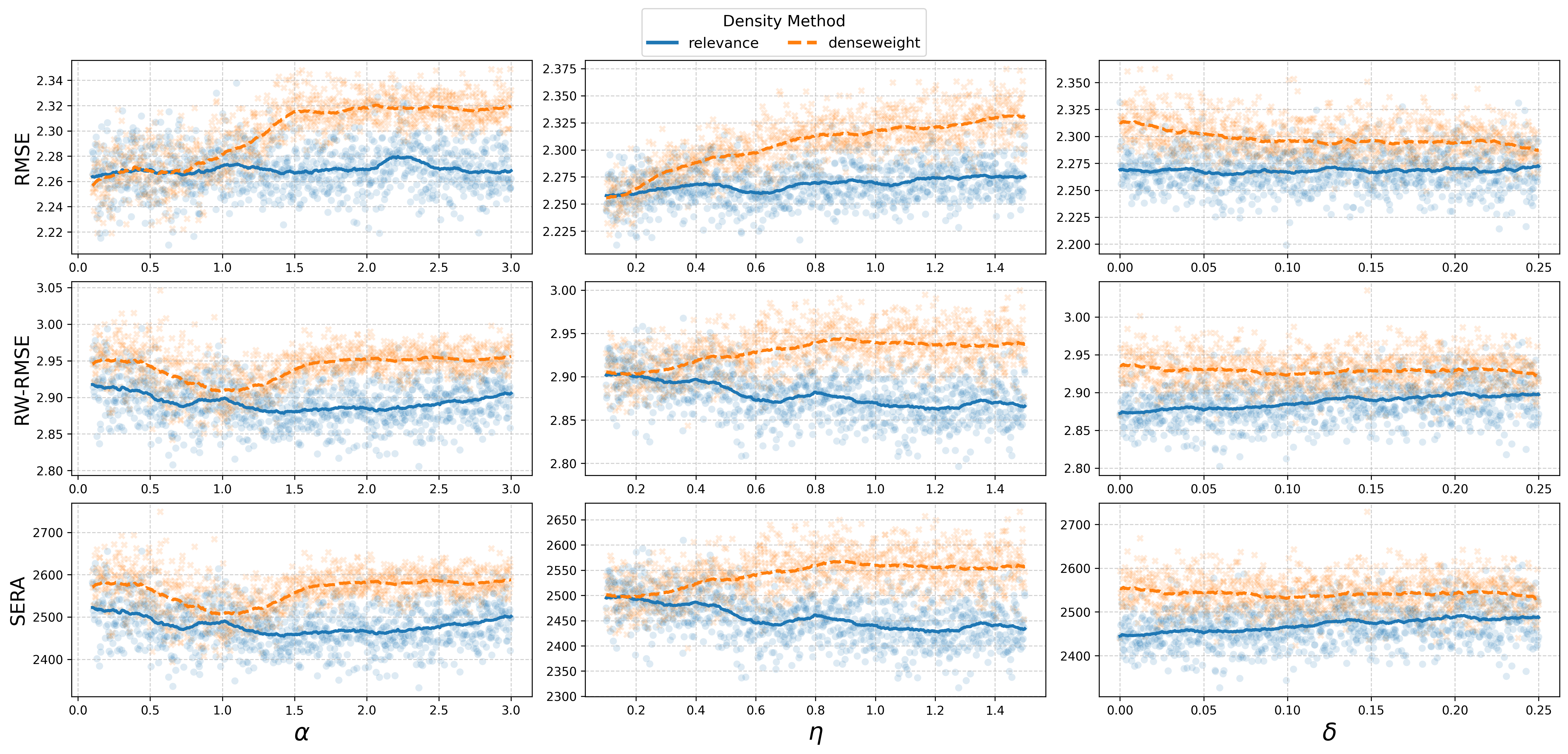}
    }
    \caption{Sensitivity analysis of CARTGen-IR with respect to the density weighting scheme ($\rho$), exponent ($\alpha$), sampling proportion ($\eta$), and noise level ($\delta$).}
    \label{fig:hyperparameter_study}
\end{figure}

We also conducted a hyperparameter sensitivity analysis of CARTGen-IR, examining the effects of the density scheme $\rho$, exponent $\alpha$, sampling proportion $\eta$, and noise level $\delta$ on performance. Results were obtained using a Random Forest model on a representative regression dataset and are summarized in Figure~\ref{fig:hyperparameter_study}.

The \emph{relevance} weighting mechanism consistently led to better performance than \emph{denseweight}. For $\alpha$, values between 1.5 and 2.5 proved most effective with \emph{relevance}, whereas $\alpha = 1$ was preferable for \emph{denseweight}. Increasing $\eta$ benefited \emph{relevance} but had the opposite effect for \emph{denseweight}. Noise had asymmetric effects as well: additional jitter improved \emph{denseweight} yet degraded \emph{relevance}.
RMSE exhibited a distinct pattern. Since it is not tailored to imbalanced regression and penalizes all deviations uniformly, emphasizing rare cases naturally leads to a slight deterioration in RMSE as $\eta$ increases. However, the other metrics, which are specifically designed for imbalanced regression, show clear improvements, indicating that the modest loss on common cases is offset by substantial gains where they matter most. This trade-off aligns with the comparative results reported earlier.

\begin{figure}[ht]
    \centering
    \resizebox{1\textwidth}{!}{
    \includegraphics{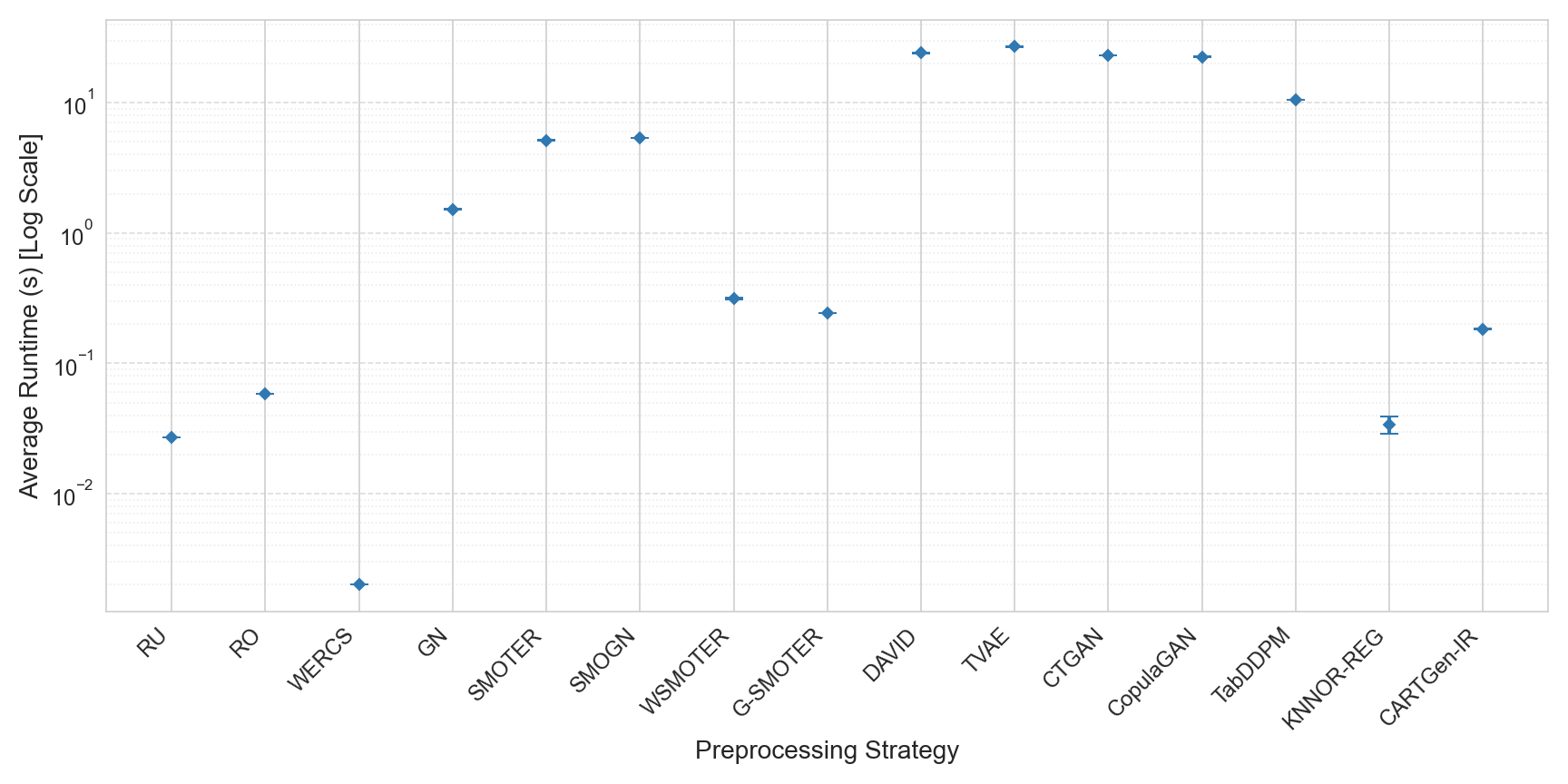}
    }
    \caption{Aggregated average runtime per data-level strategy.}
    \label{fig:runtime}
\end{figure}

To conclude our experimental study, we performed a runtime comparison across all data augmentation techniques. For fairness, we measured execution times exclusively for the data augmentation procedures under identical parallelization conditions. Figure~\ref{fig:runtime} presents the runtime values (in seconds) for each strategy, in a logaritmic scale.
CARTGen-IR stands out as one of the fastest techniques among those generating synthetic data. Sampling-based methods such as RU, RO, and WERCS exhibit the lowest runtimes, as they do not synthesize new data. Among augmentation methods that create new synthetic data, only KNNOR-REG surpasses CARTGen-IR in speed, though CARTGen-IR demonstrates lower standard deviation, indicating greater consistency.
WSMOTER and G-SMOTER have runtimes close to those of CARTGen-IR, while other SMOTER-based methods are significantly slower. Deep learning based methods are the slowest, taking on average 131 times longer to run than CARTGen-IR.

The code used in this study, along with all experimental results, is available at \href{https://github.com/antoniopedropi/SynthTabularDataGeneration-IR.git}{SynthTabularDataGeneration-IR}

\subsection{Discussion}
\label{sec:ExpDiscussion}
In response to \textbf{RQ1} and \textbf{RQ2}, 
CARTGen-IR has proven to be effective for imbalanced regression tasks. It consistently achieves strong performance across all datasets and ranks the highest overall. When compared to the leading state-of-the-art resampling strategies, CARTGen-IR either improves upon or matches these methods, particularly in balancing the focus between rare-valued and common-valued cases. For the RW-RMSE and SERA metrics, which are specifically designed to assess performance in imbalanced regression tasks, WSMOTER, KNNOR-REG, and CARTGen-IR emerge as clear winners, also demonstrating solid performance for RMSE. This indicates that CARTGen-IR generalizes well across the entire domain without compromising the overall predictive performance. Furthermore, it exhibits a significantly superior win-to-loss ratio compared to the other two methods.
Concerning \textbf{RQ3}, when analyzing the characteristics of each strategy, we can categorize the methods into sampling and augmentation techniques. Sampling techniques, due to their simplicity in implementation, result in the lowest execution times; however, they also yield inconsistent and lower-ranked scores. 
Among the methods that generate synthetic samples, KNNOR-REG is the fastest, although it performs poorly on rare target values. CARTGen-IR is the second fastest and offers a strong balance between efficiency and predictive performance.

\section{Conclusions}
\label{sec:Concl}

This work addresses the problem of imbalanced regression, where the goal is to predict rare and atypical values of a continuous target, a task that remains challenging for many learning algorithms. We introduced CARTGen-IR, a non-parametric method tailored to this setting and grounded in CART-based synthetic data generation. 
 Unlike other state-of-the-art resampling strategies for imbalanced regression, CARTGen-IR does not rely on a user-defined threshold for the continuous target. By employing a CART-based mechanism, it captures complex relationships across numerical and categorical variables and naturally handles missing data, all while remaining computationally efficient. Importantly, the synthetic data generation process itself remains transparent and interpretable, inheriting the white-box nature of decision trees.
 Furthermore, the empirical evaluation, conducted across a diverse set of benchmark datasets and  
 state-of-the-art methods, confirms the competitiveness of the proposed approach in imbalanced regression scenarios. 
 This suggests that the concept of employing a data-level strategy utilizing CART appears to be feasible. However, alternative methods can be employed to capitalize on the interactions between features and the framework laid upon by this proposal, as future work.

These findings indicate that a CART-based data-level strategy is a viable approach for imbalanced regression and point to several promising directions for future work. These include expanding the experimental evaluation to a broader range of datasets, particularly those with non-extreme rare intervals; exploring alternative decision-tree learners within the proposed framework to capture more complex feature interactions; incorporating cost-sensitive learning approaches tailored to imbalanced regression metrics such as SERA; and analysing how varying the number of synthetic samples generated per resampled instance affects overall performance.

\begin{credits}
\subsubsection{\ackname} This article is co-funded by the European Regional Development Fund (ERDF) through the Innovation and Digital Transition Programme (COMPETE 2030) under Portugal 2030 and by National Funds through the FCT - Fundação para a Ciência e a Tecnologia, I.P. (Portuguese Foundation for Science and Technology) within project FOXPM, with reference 14731 (COMPETE2030-FEDER-00923300).

\subsubsection{\discintname} The authors have no competing interests to declare that are relevant to the content of this article.
\end{credits}

%
%
%
%

\end{document}